# Aiming for Relevance

Bar Eini-Porat[1], Danny Eytan, MD, PhD[1,2], Uri Shalit, PhD[1]
[1] Technion – Israel Institute of Technology, Haifa, Israel;
[2] Rambam Medical Center, Haifa, Israel

**Abstract**
*Vital signs are crucial in intensive care units (ICUs). They are used to track the patient's state and to identify clinically significant changes. Predicting vital sign trajectories is valuable for early detection of adverse events. However, conventional machine learning metrics like RMSE often fail to capture the true clinical relevance of such predictions. We introduce novel vital sign prediction performance metrics that align with clinical contexts, focusing on deviations from clinical norms, overall trends, and trend deviations. These metrics are derived from empirical utility curves obtained in a previous study through interviews with ICU clinicians. We validate the metrics' usefulness using simulated and real clinical datasets (MIMIC and eICU). Furthermore, we employ these metrics as loss functions for neural networks, resulting in models that excel in predicting clinically significant events. This research paves the way for clinically relevant machine learning model evaluation and optimization, promising to improve ICU patient care.*

**Introduction**

Intensive care medicine is dedicated to treating critically-ill individuals who frequently contend with multiple physiological systems disturbances. Specialized healthcare professionals, including physicians and nurses, continuously track and estimate the patients' physiological state. These estimates serve as a foundation for making informed decisions regarding treatments and interventions. Moreover, given the patients' often unstable condition, their clinical state can rapidly change over minutes. Vital signs, namely basic physiological variables such as heart rate, blood pressure, respiratory rate and arterial oxygen saturation are used both to track the patient's state and to identify clinically significant changes. Predicting vital sign trajectories is valuable for early detection of possible adverse changes and to help in guiding treatment choices or intensity of monitoring[4,8]. This value stems from the capacity of such predictions to represent a patient's condition in a continuous manner, thereby assisting in the reduction of alarm fatigue and offering distinct and actionable insights to guide clinical practice. Consequently, predicting vital signs of hospitalized patients is a prevalent task in the domain of machine learning (ML) for healthcare[1,2,3,5], and clinicians from all fields express a keen interest in real-time systems for vital sign trajectory prediction[4,8].

Recent advances in deep learning now allow effective end-to-end learning of models almost directly from physiological data[2,5,9,10]. Most vital signs prediction models approach the problem as an isolated machine learning task, typically focusing on optimizing the root mean squared error (RMSE), or occasionally mean absolute percentage error (MAPE). However, ultimately such ML models are anticipated to function in conjunction with clinical experts as decision support tools; while metrics such as RMSE offer an overall assessment of performance, they assume equal importance for all predictions, which is often not the case when considering clinical utility[4].

As an example, accurately predicting a sudden drop in blood pressure typically holds much greater clinical benefit than correctly predicting no change. In the extreme, a naïve static prediction "model" which simply predicts that the value for the next time step would equal the current value, may yield respectable RMSE performance. This is primarily because patients often exhibit stability, even within an ICU setting - marked changes in state are rare, but extremely important to correctly predict. Consider the scenario where accurate predictions are made for a stable patient's vitals for 48 hours, but miss their deterioration in the final hour, thus averaging the mistake with prior successes, while exhibiting low clinical utility. Thus, exclusive reliance on RMSE or similar metrics may obscure severe errors while magnifying simple successes, resulting in an incomplete representation of the model's practical relevance. A possible solution might be transitioning to a classification problem for adverse events, thus eliminating the reliance on RMSE. However, event categorization within vital sign trajectories entails layers of complexity, by requiring the differentiation and labeling of diverse event types with arbitrary threshold values, each possessing unique clinical implications. While anomaly detection methods can effectively identify sudden surges, they may not be as applicable to longer events that persist in extreme values. Moreover, the deployment of any event-based systems (be they supervised or unsupervised) in the ICU is fraught with the risks of alarm fatigue and cognitive overloading.

In this work, we propose novel performance measures for vital sign prediction, intended to complement the classic RMSE. The goal of the new measures is to focus on dimensions of performance that align with clinical relevance as expressed by practicing clinicians. Specifically, we build on prior research by Eini-Porat et al.[4], where ICU clinicians were interviewed to assess the relevance of ML predictions of vital signs. The study identified three primary aspects of vital sign predictions that matter most in terms of utility for clinicians: vital sign values deviating from clinical normal ranges; overall vital sign trends; and deviations from those trends. These aspects were first identified via vital signs annotations tasks, then quantified by using utility ranking tasks in which clinicians ranked their interest in predictions across varying intervals, and context. This process produced aspect-specific utility curves, illustrating how perceived clinical utility changes in relation to values and trends. In this paper, we mathematically formulate utility metrics based on these empirically determined curves. We then use these metrics both as means for model evaluation, and as optimization goals for learning models which align more closely with clinical utility.

We first assess the effectiveness of the novel metrics by applying them to a simulated dataset with predetermined events that serve as ground truth to evaluate the metrics' ability to accurately capture these events, demonstrating their utility as evaluation measures for model selection. We then apply our models to clinical datasets of critically-ill patients, namely MIMIC III[6] and eICU[7] subsets, using real vital signs data to compare the clinical utility of different models.We further incorporate these metrics as loss terms for deep neural networks (DNNs). We show that models learned with the novel loss function perform better on the clinically relevant events both in simulation and real vital signs annotated by human experts. We hope these results will encourage the community to move towards more clinically-oriented modes of model evaluation and optimization.

**Methods**
We introduce three utility-based costs which correspond to clinically significant aspects of a vital signs' behavior[4]: (i) exceeding clinical normal range, (ii) overall trend, and (iii) trend deviations. We use these utility costs in two ways: The first is as evaluation metrics which allow different perspectives on the performance of any given prediction model. The second is as optimization targets, specifically as loss terms for DNNs, aiming to promote model performance which aligns more closely with each of these three aspects of clinical utility.

**Design**
**Utility Costs: Formulation**
According to clinicians, perceived clinical utility is directly related to clinical severity and surprise, i.e. how unexpected a trajectory is. High or low absolute values are important as they describe clinically severe states - a patient with a predicted heart rate of 180 BPM is probably in danger. General trend hints at clinical severity even if the absolute values are not severe just yet - the clinical staff could be missing a deterioration. As for sudden deviations from the trend, a sudden change could be a symptom of a clinical event or simply less likely to be predicted by the clinical staff and therefore novel. Here we mathematically formulate utility curves that capture each aspect, in the form of functions that measure the utility cost of giving a certain prediction with respect to the ground truth.

Normal Range
As found by Eini-Porat et al.[4], measurements' clinical importance follows a sigmoidal pattern around the norm thresholds, with vital sign values perceived as more important as they exceed the clinical norm. However, at some extreme values further deviations from norm do not lead to a change in perceived clinical importance since the patient is likely already experiencing an event demanding attention. In the prevalent case where there are both lower and upper bounds for the acceptable range of a vital sign, the curve would manifest as two mirror sigmoid functions; see for example Figure 1 for heartrate[4].

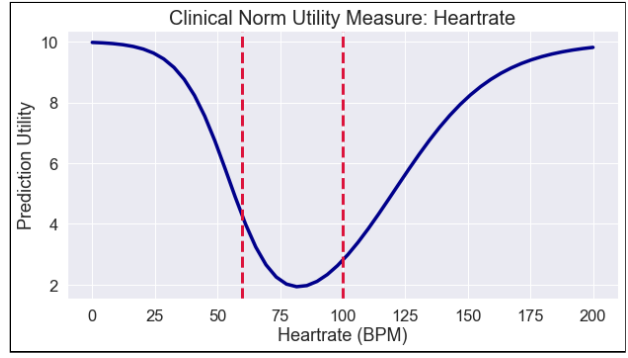

**Figure 1.** Normal range utility: heart rate example. The values between the red lines are the clinically normal range. The blue curve is the calculated normal range utility curve.

To capture this behavior, we define the normal range utility cost as follows:

**Equation (1), normal range utility cost:**

$$U_r(y_t, \hat{y}_t) = \left| \left( max\left(\frac{L}{1+e^{k_h \cdot (y_t-h)}}, 0\right) + max\left(\frac{L}{1+e^{k_l \cdot (y_t-l)}}, 0\right) \right) - \left( max\left(\frac{L}{1+e^{k_h \cdot (\hat{y}_t-h)}}, 0\right) + max\left(\frac{L}{1+e^{k_l \cdot (\hat{y}_t-l)}}, 0\right) \right) \right|,$$

where $y_t$ represents the true value and $\hat{y}_t$ prediction at time $t$, and $U_r(y_t, \hat{y}_t)$ is the utility cost of predicting $\hat{y}_t$ while the true signal is $y_t$. The terms represents a two-sided sigmoid, for example the blue curve in Figure 1. The parameters $L, k_l, k_h, h, l$ correspond to low and high threshold sigmoids: $L, k_h$, and $k_l$ indicate the magnitude and shape of these sigmoid curves, while $l, h$ refer to the range thresholds. These parameters enable adaptation to various vital signs and the portrayal of any of the clinical norm utility curves identified by Eini-Porat et al[4]. Looking at Figure 1, we can see that, all else being equal, $U_r(y_t, \hat{y}_t)$ will attain larger values when the prediction is for a normal range value but the outcome is out of normal range, or vice versa.

Trends
To establish the utility measures for overall trend and trend deviation, we start with some definitions: (1) The expected trend $Y'_{t-n:t}$ refers to the slope of the vital over the preceding $n$ steps, (2) Actual trend $Y_{t-n:t+m}$ is the true signal slope over the past $n$ and extending to the subsequent $m$ intervals and (3) Predicted trend $\hat{Y}_{t-n:t+m}$ is the projected slope for the next $m$ steps, based on the previous $n$ steps as estimated by a ML model (see Figure 2). We note that the size of $n$ and $m$ may vary based on frequency of the vital sign measurement. For example, a high frequency vital sign is more susceptible to noise compared to low frequency ones, thereby introducing noise to these measures. Consequently in such cases larger values of $n$ and $m$ might be necessary to mitigate noise.

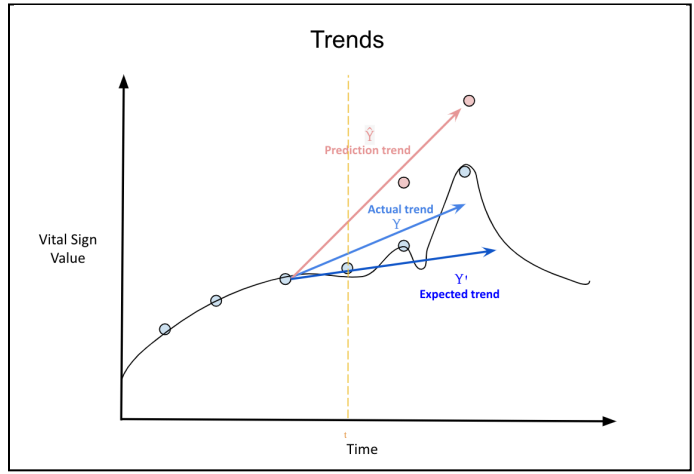

**Figure 2.** Illustration of trend terms. Blue points represent the true vital sign measurements, and pink are predictions. The pink arrow represents the predicted trend $\hat{Y}_t$, the light blue arrow represents the actual trend $Y_{t-n:t+m}$, and blue arrow is the expected trend $Y'_{t-n:t}$.

The Trend utility error considers the disparity between actual and predicted trends, using a quadratic loss - a slight trend change could be benign, but a sharper one might indicate an underlying event unfolding. Indeed, clinicians' utility does not increase linearly; it rises rapidly as the trend becomes steeper[4]. We note that trend prediction costs might not be symmetric, for example a drop in blood pressure could be more dangerous than an equal-sized surge. The following formulation allows for such asymmetries through the non-negative weighting terms $w_l, w_h$.

**Equation (2), trend utility cost:**
$$U_{\text{tr}}(Y_{t-n:t+m}, \hat{Y}_{t-n:t+m}) = max(\hat{Y}_{t-n:t+m} - Y_{t-n:t+m}, 0)^2 w_l + max(Y_{t-n:t+m} - \hat{Y}_{t-n:t+m}, 0)^2 \cdot w_h$$

While the trend and clinical norm aspects were captured by utility curves in Eini-Porat et al.'s study, the trend deviation aspect was not directly quantified. Yet its significance to clinicians is paramount, as they frequently mentioned it in various scenarios. We propose a trend deviation utility measure which is the trend prediction errors, but weighted by the difference of expected (by linear extrapolation) and actual trends, or *"how surprising is that trend?"*.

**Equation (3), trend deviation utility cost:**
$$U_{\text{td}}(Y_{t-n:t+m}, \hat{Y}_{t-n:t+m}, y_t, \hat{y}_t) = (Y'_{t-n:t} - Y_{t-n:t+m})^2 \cdot |y_t - \hat{y}_t|$$

**Utility Costs: Evaluation and Optimization**

Given a time-series of vital signs and vital sign predictions, each of the three utility costs $U_r, U_t, U_{td}$ can be readily calculated and averaged over time and over signals. We use these averages both as metrics to evaluate the utility costs of given models (*evaluation*), and as terms in loss functions for deep neural networks trained in mini-batches over the dataset (*optimization*).

For *evaluation*, we train multiple ML models (see Models) to predict the next measurement using three datasets. First, a simulated vital signs dataset is used, where input data from the previous three time steps is utilized as input. Subsequently, we use two clinical ICU datasets with the objective of predicting mean blood pressure in the next 30-minute interval within a 24-hour period. Our primary objective in evaluating the simulated dataset is to validate the newly proposed utility costs. With the ground truth of event locations, we can assess the utility costs' ability to accurately capture them. The evaluation on clinical datasets, on the other hand, represents a real-world use case where we can measure the added value of these metrics in the model selection process.

For *optimization*, we aim to actively improve the utility-based costs of a model. We train DNNs (LSTM) with loss functions incorporating terms based on each utility aspect, for example: $loss = MSE\ loss + \lambda \cdot \frac{1}{n}\sum U_{\text{tr}}$, to examine whether they could promote the corresponding utility costs. The combination with the standard squared loss is necessary as some of loss terms lead to objective functions that deviate from target signal. Similarly to the evaluation phase, we train to predict the next measurement over the three datasets. Finally, we train a mixed utility model using a training regimen that integrates loss terms of different aspects. We assess the performance of this model on a small annotated eICU subset, which includes mixed event annotations.

**Models and training**

We train different ML models for both phases, including Linear regression, Ridge regression, Bayesian ridge, XGBoost regressor (all using *scikit-learn[11]*) and LSTMs using *pytorch*[12]. Results are reported for 5-fold cross validation. LSTM models architecture comprises an encoding layer, a 2-layer LSTM structure and a fully connected final layer. In the optimization phase the proposed utility-based errors find application as loss terms within Deep Neural Networks (DNNs). These measures, however, often prioritize behaviors that compete with accuracy. The intricate balance between accuracy and these measures necessitates a warm start training regime. Initial training prioritizes accuracy through MSE optimization, succeeded by a subsequent fine-tuning phase involving utility loss terms, progressively escalating the utility loss coefficient $\lambda$. In the optimization of multiple cost terms, a new utility cost was introduced only after the stabilization of the previous cost components. The $\lambda$ hyperparameter was selected to balance loss terms while maintaining the same scale of values across the utility metrics.

**Datasets**

**Simulation**

In this work we challenge the heavy reliance on standard evolution metrics and propose new measures to complement them. We first wish to show that these measures work as intended in a controlled situation. Thus, we generate a synthetic vital signs simulation where we artificially introduce several types of events on top of a continuous signal. We then test the new metrics compared to RMSE over the time windows corresponding to the introduced events.

The simulation consists of 5 variables describing the base periodic process. These variables are constructed as follows for patient $i$ at time $t$:

$a_i \sim \text{uniform}(18, 80)$ is a static "demographic" parameter.
$g_i \sim bin(0.5)$ is a static "demographic" parameter.

$s_{1,i}(t) = c_{12} \cdot a_i + c_{12} \cdot g_i + c_{13} \cdot sin(\alpha_1 t) - c_{13} \cdot sin(\alpha_1 t) + \sigma_{1,i}(t)$
$s_{2,i}(t) = c_{22} \cdot a_i + c_{22} \cdot g_i + c_{23} \cdot sin(\alpha_2 t) - c_{24} \cdot sin(\alpha_2 t) + \sigma_{2,i}(t)$
$s_{3,i}(t) = c_{31} \cdot sin(\alpha_3 t) + c_{32} \cdot s_{1,i}(t-1) + c_{33} \cdot s_{2,i}(t-2) + \sigma_{3,i}(t)$
$\sigma_{k,i}(t) \sim \mathcal{N}(0,1)$ i.i.d. for all $k, i, t$.

The variables $a_i$ and $g_i$ set a base level for the signals, and the time series is a linear combination of sinus functions and noise; $\alpha_i$ and $c_{ij}$ determine the frequency of the periodic changes and their magnitude. The target signal is $s_3$, and it depends on previous measurements of $s_1, s_2$. These baseline variables are embedded with different types of events: rare sudden events and evolving trend events. These events are generated separately and added to the relevant signals. For rare sudden events, we introduce events of two types. The first is a sudden drop with fixed size 7 in $s_2$ which leads a surge in the next time step of $s_3$. The drop occurs with small probability $p_1 = 7 \cdot 10^{-4}$ per time step, given that the drop occurred a surge follows with probability $p_{31} = 0.75$ with random duration $d_1 \sim \text{Uniform}[0, 10]$. The second rare event is a surge in $s_2$ which results in a surge in $s_3$ two time steps later of the same size. The probability $p_2 = 5 \cdot 10^{-4}$ is with a random duration $d_2 \sim \text{Uniform}[0, 10]$. with probability $p_{32} = 0.6$. Finally, the trend event is added to $s_3$ with probability $p_{33} = 1 \cdot 10^{-4}$ per time step and a random duration $d_3 \sim \text{Unifrom}[0, 10]$. We note that for the evaluation we considered an additional event type, range events, which are the cases where measurements were further than 2 standard deviations from the mean. We generate a simulated dataset with 1000 patients each with 500 time steps, with an average of 1.5 events per patient.

**Clinical Datasets**
Our goal is to predict vital sign trajectories for hospitalized patients. To this end, we used two open ICU databases: MIMIC-III and eICU, focusing on heart rate and blood pressure signals. The MIMIC-III Waveform Database[6] Matched Subset contains thousands of waveform records of ICU patients hospitalized in critical care units of the Beth Israel Deaconess Medical Center between 2001 and 2012. This database is a subset of the MIMIC-III Waveform Database, representing those records matched clinical records are available in the MIMIC-III Clinical Database containing electronic health records. Since the prediction target is primarily next step heartrate prediction during a 24 hours window, only records with heartrate waveforms recordings with durations that exceeded 12 hours were selected yielding 284 patient records. Features extracted from MIMIC-III Clinical Database and aggregated to half an hour window in accordance with ICU clinicians desired prediction time scales[4]. The feature set includes pervious aggregated heartrate mean and standard deviation, mean atrial pressure and standard deviation from the waveform dataset, merge with clinical dataset features age, gender, weight, height, respiratory rate, systolic and diastolic blood pressure, temperature, SpO2 and a "pain present" indicator.

The eICU Collaborative Research Database[7] is a large dataset containing over 200,000 records gathered from hundreds critical care units throughout the continental United States. The data in the collaborative database covers patients who were admitted to critical care units in 2014 and 2015. Due to compute limitations 1,000 patient admissions were sampled out of this dataset. The features we use consist of previous heart rate, blood pressure, SpO2, respiration rate, temperature aggregated to 30 minutes steps and age and gender. To reduce noise, records with at least 50% of the SpO2 and respiration rate data present during the times when heartrate measurement is present were selected, leaving 389 records.

Finally, as an external validation we collected vital signs annotations for small subset of 45 signals of eICU subset by 4 different clinicians (3 residents and 1 nurse). Annotators were instructed to mark segment of the heartrate signal which they would consider important to predict. They were compensated with coffee vouchers worth ~10$. The annotation task was approved by our institutional IRB.

**Main outcome measurement**

To validate the proposed metrics, we compute the proposed utility measures and contrast them with the calculated RMSE over the various types of simulation events. Then, we assess whether these metrics provided any additional value compared to the overall RMSE-based evaluation across all datasets.

In the optimization phase, we examine whether training with a given utility loss leads to different model behavior, and whether it leads to lower utility costs of the model when evaluated on held-out data. This examination allows us to evaluate whether training to improve utility is at all possible. We note that in real-world scenarios, these utility costs are likely to be the sole quantitative measure to check for training success. Finally, we assess the RMSE of a baseline LSTM model and the Mixed model (an LSTM optimizing all utilities jointly) over time windows corresponding to clinical events as annotated by clinicians in the eICU dataset.

**Results**

**Simulation dataset**

**Evaluation.** Figure 3 illustrates the contrast between the costs assigned by RMSE and by $U_{td}$ with respect to identical model behavior. While both assign penalties for missing the drop event around $t = 70$, RMSE is noisier and assigns penalties to small prediction mismatches throughout the signal.

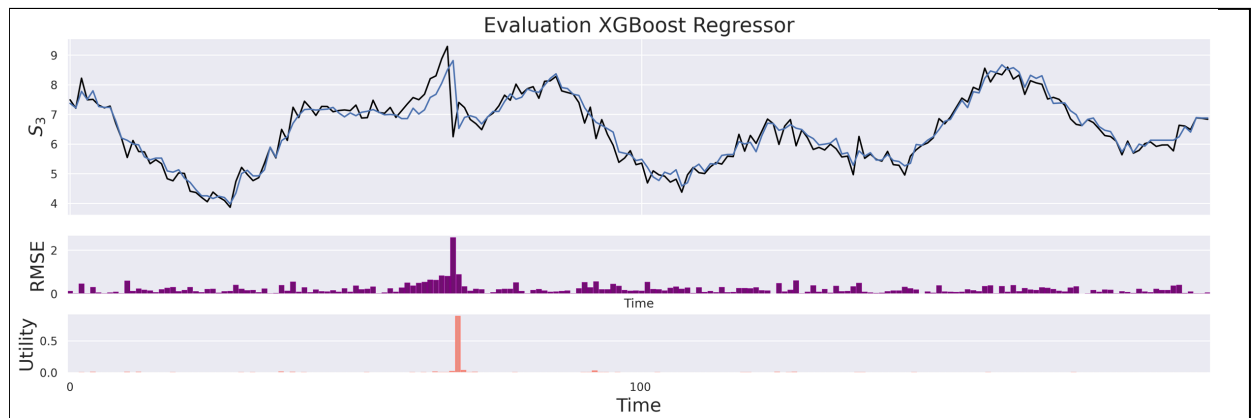

**Figure 3.** Simulation data: The top panel shows true values $y_t$ (black) against XGBoost regressor step predictions (blue) for the simulated signal $s_3$. The middle and bottom panels display RMSE and $U_{td}$ cost, respectively. The $U_{td}$ cost is only sensitive to the sharp mismatch at the "drop" event.

At the macro level, we show in Figure 4 that over the three event types - range, trend, and trend deviation - models with low utility cost have low RMSE when measured specifically over the time window of the corresponding event. Thus, selecting models based on utility costs would lead to models which are more accurate around specific types events, possibly at the expense of higher overall RMSE.

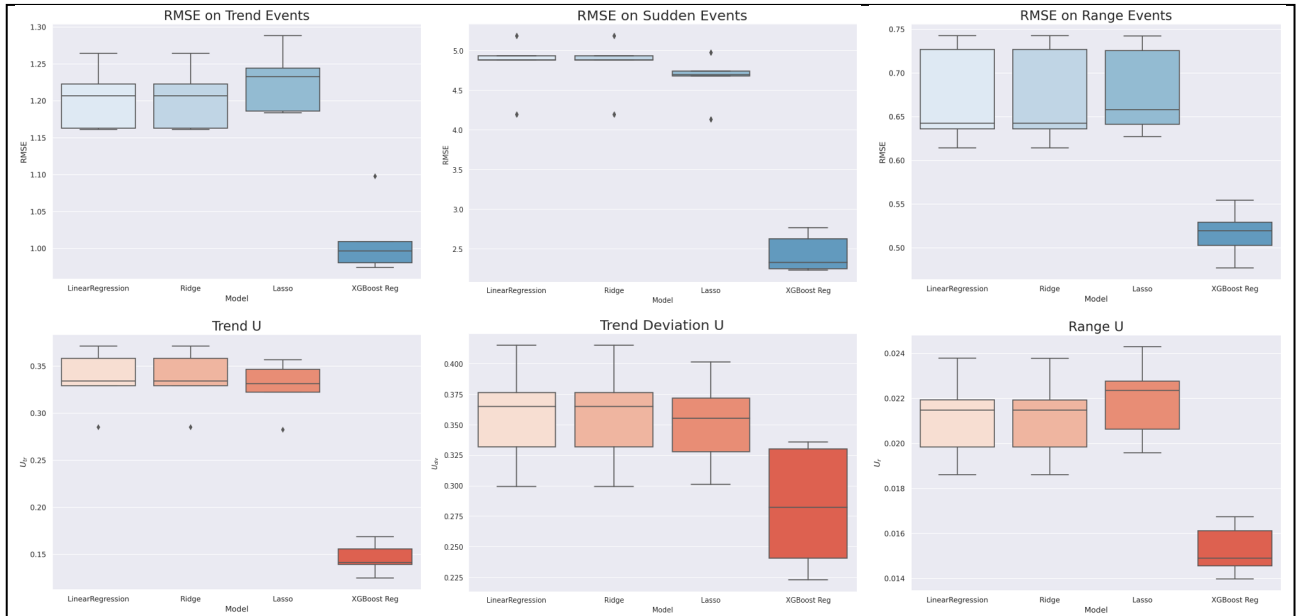

**Figure 4.** Simulation data: The top row presents RMSE over different types of events per model. On the bottom is the corresponding utility cost tested for the same models.

**Optimization.** In Figure 5 we see the ground truth signal as well as predictions made by two LSTM models: one trained with standard squared loss, and one trained with an additional trend deviation utility cost term. We see that the utility term encourages the model to be more precise over the surge event near $t = 100$, at the expense of slightly lower accuracy diffused over the rest of the signal, including a more minor surge around $t = 50$.

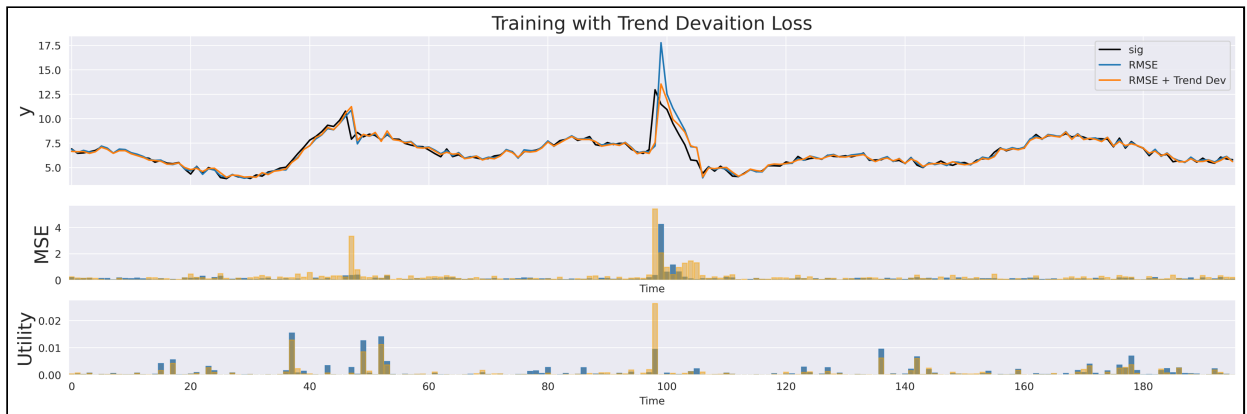

**Figure 5.** LSTM model trained with RMSE loss for 6000 epochs (blue) vs. LSTM model trained with $U_{td}$ component (orange) for 5000 epochs. With less training, the utility oriented model proves to be more accurate on sudden events.

In Figure 6, we show the performance of LSTM models trained with different utility costs, including a baseline model with only the squared loss. We examine performance in terms of RMSE over the intervals containing the three types of rare events we introduced into the signals, as well as performance in terms of the utility costs. We see the correlation between RMSE on the events and the corresponding utility metrics. We further see that models trained with the relevant utility cost indeed perform better with respect to that utility.

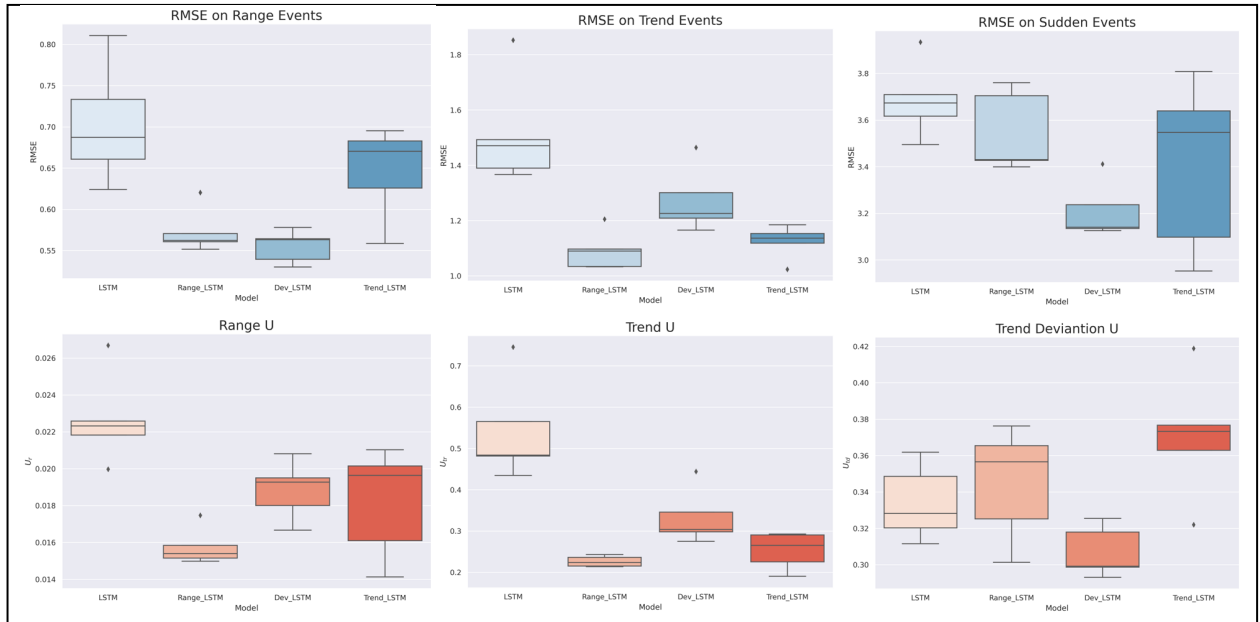

**Figure 6.** Simulation data: Performance of LSTMs trained with utility loss, in terms of RMSE over events of different types, as well as the corresponding utility metrics. Baseline LSTM is marked as 'LSTM', the Clinical range trained model is marked as 'Range_LSTM', the model trained for overall trend as 'LSTM_Trend' and the model trained for trend deviations is 'LSTM_Dev'.

### Clinical Datasets

**Evaluation.** Figure 7 presents the performance on MIMIC and eICU blood pressure prediction task of four standard machine learning models. The results are presented in terms of RMSE and in terms of the three utility costs we introduce; all results are on held-out data. Using the standard RMSE loss, the XGBoost model demonstrates superior performance, whereas the LSTM model lags notably. However, a different picture emerges when we look at the other utility costs: the LSTM model significantly outperforms the other models for trend and trend deviation, while still underperforming for the range utility cost. . These distinct behavioral patterns of the LSTM models underscore the necessity for a context-sensitive ranking, influenced by deployment considerations. In the process of model selection, a superficial reliance on RMSE might lead to dismissing LSTM, but the comprehensive range of measures suggests that LSTM should have merited consideration and potentially even selection, contingent upon the specific objectives of the task. We obtained similar results in the model selection for a heartrate prediction tasks across both datasets.

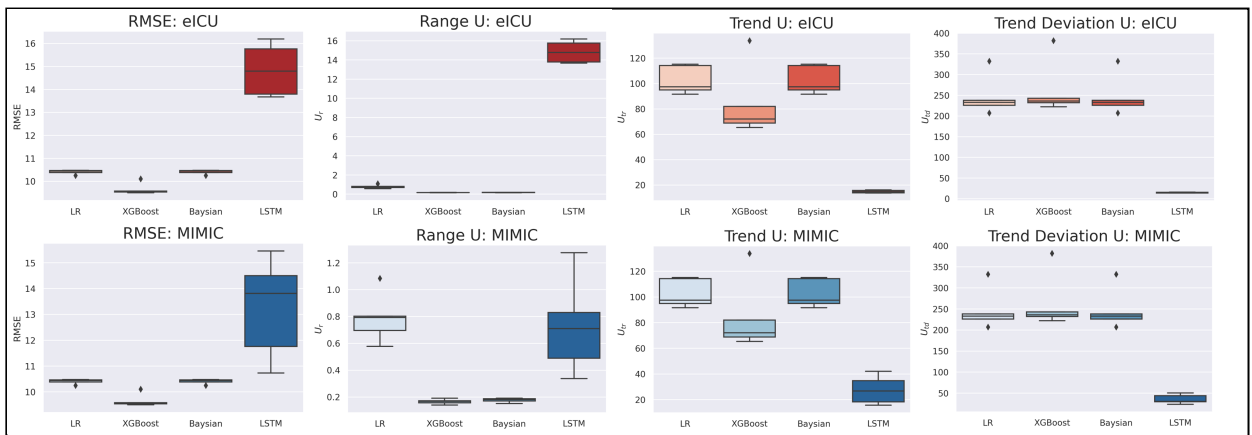

**Figure 7.** eICU and MIMIC subsets blood pressure prediction task. The performance according to RMSE and each of the three utility costs is shown. The top row depicts these measures per model for the eICU dataset, and the bottom row depicts the same measures for the MIMIC dataset.

**Optimization.**
When optimizing over the MIMIC subset we observed that training led to a decrease in all utility losses. However, we also observed some mild instability in the range utility cost, where a certain split in the 5-fold validation failed to achieve an adequate baseline performance. While a reduction in utility loss is achieved over eICU subset, the improvements seem less significant, possibly due to the high rate of missingness, hindering the learning process.

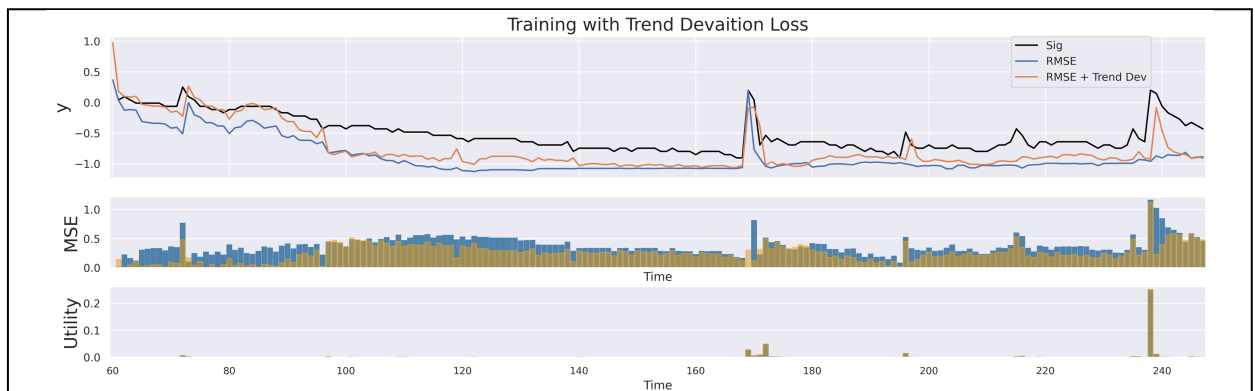

**Figure 8.** Heartrate prediction over the eICU subset. Baseline model trained with RMSE loss (blue) vs. LSTM model trained additionally with $U_{td}$ component (orange). The utility-optimized model captures more events. In the bottom row, the blue and orange bars overlap.

In Figure 8 we present the performance of an LSTM trained with standard squared loss vs. an LSTM trained with an addition trend deviation term, on the task of predicting mean heartrate in the eICU dataset. We see that both models have some errors with respect to the true signal, but the trend deviation model captures several of the spiking events more accurately.

Finally, we compare the performance of a standard LSTM model trained with square loss, with an LSTM model trained with a combination of all three utility measures, which we call a Mixed model. We use the annotated dataset where clinicians marked heartrate segments which they deemed clinically significant, and which could correspond to each of three clinical utility aspects we outlined, or indeed any other aspect that came to their mind, marking overall 94 events over 45 patients with average length of 15 steps. The held-out RMSE over the entire signal for the baseline model is 1.205 while the corresponding RMSE of the Mixed model is only slight better at 1.167. However, the Mixed model had a greater advantage in RMSE taken only over the annotated events 1.276 compared to 1.485 for the baseline. A paired t-test over the 94 events gives a statistic of -1.96 with a p-value of 0.056. In Figure 9 we present a histogram of the differences in RMSE over the annotated events.

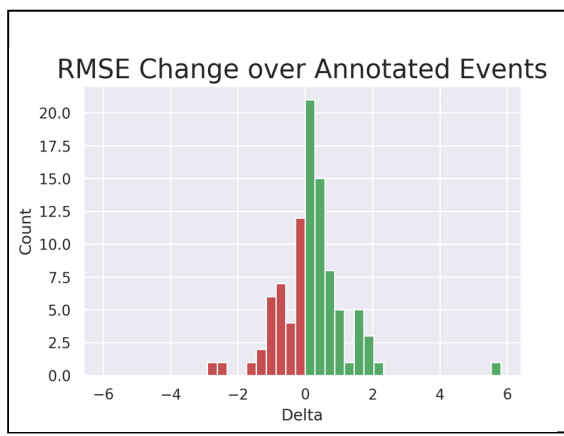

**Figure 9**. Difference in RMSE over annotated events between vanilla LSTM model and Mixed model trained with utility costs. The task is heartrate prediction in the eICU dataset.

**Conclusions**
In this work we argue that for ML-based vital sign prediction models to be truly useful in practice, their evaluation should consider clinically relevant aspects of the prediction problem. We thus introduce new utility measures of model performance whose goal is to complement standard metrics such as RMSE. The idea behind these new measures is the recognition that a prediction's meaning is context-dependent, encompassing both clinical and temporal contexts.

The newly introduced metrics are applicable to model selection and model optimization whenever vital sign prediction is the target. This allows overcoming the problem of models which are accurate on average over the entire time-series, but in fact perform poorly with respect to clinical goals. The defined utility costs provide a way to influence the model's emphasis, for example prioritizing trend prediction or deviations from clinical norm.

During the optimization phase, while we anticipated a more distinct trade-off between utility costs and RMSE, we found it noteworthy that overall RMSE exhibited only a slight reduction following training with any of the utility measures in both datasets. This phenomenon could potentially be attributed to the utility loss measures serving as regularization factors within the training process. An inherent limitation of a work calling for new evaluation metrics is how can one evaluate novel metrics, given that they are meant to highlight shortcomings of existing metrics. We attempted to overcome this limitation using both a simulated dataset and annotations from clinicians over real vital sign signals, showing that training a model with the proposed costs possibly leads to smaller error on annotated events thus indicating its clinical usefulness.

It is our hope that the introduction of these new metrics will enhance vital signs prediction models, given their perceived utility by ICU clinicians[4]. Although our work primarily focuses on vital sign predictions, the next step involves exploring the incorporation of these metrics into a larger, more comprehensive model for patient state and patient deterioration. Indeed, while the proposed utility costs mark a step forward in this direction, it is worth noting that the incorporation of context could be extended further. The interplay between vital signs could also offer insights into clinical severity and surprise within each of the three aspects: clinical norm, overall trend, and trend deviation. Additionally, the integration of general EHR data and clinical knowledge bases is a possibility – for instance, a drop in heart rate following prescribed medication may be expected and non-concerning. Another future direction could focus on the implications of deploying models learned with different utility costs in the ICU environment, and how are they perceived by practicing clinicians in the field. Finally, the newly introduced utility costs could be found in the field of interpretability. Such scores could be utilized for detecting samples with high clinical value for a review of the model performance.

**Acknowledgements**
We gratefully acknowledge funding from VATAT – data science and the Israeli Science Foundation grant 2456/23. We further wish to thank the support of the Technion-Rambam initiative in AI in medicine and the Technion Data Science initiative.